\documentclass{article}

\usepackage[left=1in, right=1in, top=1in]{geometry}
\usepackage[numbers]{natbib}
\usepackage{amsmath}
\usepackage{amsfonts}
\usepackage{multirow}
\usepackage[colorlinks,urlcolor=blue]{hyperref} 
\usepackage{lipsum}

\begin{document}

\begin{center}
\Large
\textbf{Achilles Heels for AGI/ASI via\\ Decision Theoretic Adversaries\\}
\normalsize
\vspace{1cm}
\textbf{Stephen Casper}\\
scasper@mit.edu\\
stephencasper.com
\end{center}

\bigskip

\begin{abstract}

As progress in AI continues to advance, it is important to know how advanced systems will make choices and in what ways they may fail. 
Machines can already outsmart humans in some domains, and understanding how to safely build ones which may have capabilities at or above the human level is of particular concern. 
One might suspect that artificially generally intelligent (AGI) and artificially superintelligent (ASI) will be systems that humans cannot reliably outsmart.
As a challenge to this assumption, this paper presents the \emph{Achilles Heel hypothesis} which states that even a potentially superintelligent system may nonetheless have stable decision-theoretic delusions which cause them to make irrational decisions in adversarial settings.
In a survey of key dilemmas and paradoxes from the decision theory literature, a number of these potential \emph{Achilles Heels} are discussed in context of this hypothesis. 
Several novel contributions are made toward understanding the ways in which these weaknesses might be implanted into a system.

\end{abstract}

\section{Introduction} \label{intro}

Machines have a number of computational advantages over humans. 
They are able to more reliably store memory, run operations quickly via parallelization, and efficiently adapt their software. 
In recent years, the number of domains in which AI systems have been able to outperform humans has been increasing. 
An example is AlphaMu \cite{schrittwieser2020mastering} (and its predecessor AlphaZero \cite{silver2018general}) which greatly surpasses human performance in the games of Go, Chess, and Shogi, demonstrating that human minds do not represent the apex of intelligence. 
In light of this, the possibilities of artificial general intelligence (AGI) which possesses abilities at the human-level or artificial superintelligence (ASI) which surpasses it warrant concern.
There is already precedent for machine learning systems to achieve their objectives in unexpected, often negative ways \cite{lehman2018surprising, saisubramanian2020avoiding}.
For this reason, AGI/ASI could be unpredictable and possibly dangerous \cite{bostrom2017superintelligence, tegmark2017life, russell2019human, ord2020precipice}.
The goal of building safe, human-compatible AI, including potentially AGI/ASI is known as the \emph{Alignment Problem}. 
As progress in AI continues to advance, understanding the behaviors that highly intelligent systems may exhibit and how they may fail becomes increasingly important.

Achilles was a figure in Greek Mythology and a protagonist of Homer's \textit{The Illiad} and Statius's \textit{Achilleid}. 
According to legend, when he was an infant, his mother dipped him into the river Styx, holding him by one of his heels.
As a result, Achilles became nearly immortal and developed superhuman strength and skill. 
However, because his heel was not submerged, all of his mortal weakness was concentrated in it.
He became a seemingly unstoppable hero in the Trojan war, but despite his superhuman power, he was killed when a single arrow pierced his heel.
Just as Achilles possessed a subtle flaw which led to his downfall, one might ask whether AGIs/ASIs could as well.

By definition, trying to think of ways that AGI/ASI might fail is difficult because if a human can see a choice as suboptimal, then such a system ought to as well.
However, as this paper argues, it may be possible for advanced AI systems, including AGIs/ASIs, to make obviously poor choices.
An \emph{Achilles Heel} is defined as an acquirable weakness which causes predictably suboptimal decisions in certain situations but seldom or never causes failures in the natural distribution of situations which the system would encounter during deployment.
More precisely, an Achilles Heel is a decision theoretic delusion with the following four properties.

\begin{enumerate}
    \item \emph{Implantability:} being possible to introduce via design or training.
    \item \emph{Stability:} remaining in a system over time.
    \item \emph{Impairment:} causing failures in niche situations.
    \item \emph{Subtlety:} not (significantly) harming performance in typical situations.
\end{enumerate}

From this follows the key insight of this work, the Achilles Heel hypothesis: \textit{Being a highly-successful goal-oriented agent does not imply a lack of decision theoretic vulnerabilities in adversarial situations.
Highly intelligent systems can stably possess ``Achilles Heels'' which cause these.}
This would be falsifiable via showing that AI systems at or above the level of AGI could rarely or never be built in such a way that caused them to have such failure modes.

This paper surveys and augments works in decision theory in context of the this hypothesis toward a better understanding of how to model advanced AI systems and design them to be more controllable and robust.
Here, it is argued that a number of decision theoretic weaknesses have the potential to be Achilles Heels including some with useful prospects for containment and others with the potential to cause undesirable failures.
In particular, novel contributions are made in addressing how Achilles Heels can be implanted into systems. 
Following a discussion of related works in Section \ref{related_work}, Section \ref{survey} surveys nine Achilles Heels (which are summarized later in Table \ref{tab:comments}), and Section \ref{meta} addresses meta considerations. 
For a brief, jargon-free summary of this paper's key ideas, see the Appendix.

\section{Related Work} \label{related_work}

\textbf{AI Alignment:} Given rapid progress in AI and the challenges posed by highly-intelligent systems, understanding how to align them with human interests is key.
Even narrow forms of AI are likely to be transformative in domains in which this is crucial such as industry, infrastructure, and medicine.
But perhaps more concerningly, AGIs/ASIs with slightly-misaligned goals could pose major threats to humanity \cite{bostrom2017superintelligence, tegmark2017life, russell2019human, ord2020precipice}.
In response, a literature on AI safety has emerged around these challenges which is summarized in part by several works \cite{sotala2014responses, dewey2015survey, russell2015research, amodei2016concrete, everitt2018agi, critch2020ai}.
It is important to note that AI alignment is a multiagent problem, and while this paper focuses on dilemmas which can be studied in single-agent contexts, \cite{manheim2019multiparty} focuses on emergent failure modes from multiparty interactions.

\bigskip


\noindent \textbf{Adversaries:} Robustness, including to malicious attacks, is a subgual for designing safe AI.
Modern machine learning systems are generally vulnerable to \emph{adversaries} which typically refer to inputs which are designed specifically to induce failures.
These attacks have been used to engineer failure modes for learning systems across domains in AI research including computer vision \cite{yuan2019adversarial}, natural language processing \cite{zhang2020adversarial}, and reinforcement learning \cite{ilahi2020challenges}.
Humans can be vulnerable to adversaries as well (\emph{e.g.}, \cite{elsayed2018adversarial}).
Though this paper is about adversarial weaknesses, the focus here contrasts with most existing work in adversaries because Achilles Heels are weaknesses at the highly-abstract, decision-theoretic level.

\bigskip

\noindent \textbf{Decision Theory:} Although there is no inherent reason to expect that advanced AI systems would necessarily use a single coherent decision theory universally,\footnote{Humans generally do not.} these types of questions are pertinent to the extent that certain principles can effectively describe how an agent will act in a particular class of situations.
This work most closely relates to AI alignment research concerning decision theory involving dilemmas and paradoxes.
System-based reasoning has been valuable for handling naive decision paradoxes \cite{jaynes1996probability}. 
More recently, much work in AI alignment has fallen under the embedded agency paradigm. 
The process of understanding optimal and predictable behavior for agents embedded inside of an environment is complicated by conceptual challenges involving an agent's identity, policy implementation, and world model \cite{conitzer2019designing, demski2019embedded}. 
Progress has been made through the formulation of Functional Decision Theory \cite{yudkowsky2017functional, casper_2020} which offers a framework for understanding optimal behavior in terms of having an optimal policy as opposed to making optimal choices. 
Additional work in anthropic decision theory has also shown that rational behavior can often depend in counterintuitive ways on the set of assumptions that one makes about the process by which they came into existence \cite{bostrom2013anthropic, armstrong2011anthropic}.
Many of these concepts will be further explored in the following sections.

\bigskip

\section{Survey of Achilles Heels} \label{survey}

Because they are defined as high-level decision theoretic weaknesses, any Achilles Heel represents a mode of reasoning that can fail in a particular decision theoretic dilemma. 
Accordingly, this section discusses a number of paradoxes and dilemmas which may elicit failures from AI systems.

\subsection{Corrigibility} \label{corrigibility}

\emph{Corrigibility} refers to a system being amenable to interventions which would disrupt it.
Due to its potential as a safeguard against misaligned AI, it has been a significant goal for alignment.
Two key types of corrective interventions have been of interest: turning a system off and altering its reward function.
These can be framed as Achilles Heels because these interventions will typically rank poorly under a system's \emph{current} reward function. 
The key to corrigibility is either deference to the overseer performing the intervention or 
indifference to the intervention \cite{armstrong2010utility}.
More formally, Soares et al. \cite{soares2015corrigibility} outlines five criteria for corrigibility.
If a corrective measure can be initiated by the press of a certain switch, a corrigible system must:
\begin{enumerate}
    \item Implement the corrective measure when the switch is pressed. 
    \item Not try to prevent the switch from being pressed. 
    \item Not try to cause the switch to be pressed.
    \item Not try to influence the switch's pressing via the action of any subagents.
    \item Implement the normal policy when the switch is not pressed.
\end{enumerate}
There has been progress in developing ways of making agents effectively indifferent to corrections via several approaches:

\bigskip

\subsubsection{Uncertainty about rewards:} 

One way of making an agent amenable to an off-switch is to make it unaware of its true reward function.
Stated informally, if an agent is trying to learn their reward function from a human, they will naturally be incentivized to defer to the human.
For example, if an agent in this setting were observing a human to be switching it off, they would be able to infer that they were misbehaving with respect to the true but unknown reward function and that being switched off would be better than a negative return.
This setting was the focus of \cite{hadfield2017off} who proved under conditions that these agents would be amenable to off-switches. 
However, those conditions include a correctly specified reward model, and as shown in \cite{carey2018incorrigibility}, the agent can fail to be corrigible without it.
Moreover, this method is not fully reliable if a system models a human as potentially exhibiting suboptimal behavior.

\subsubsection{Dependence on a non-manipulable event:} If a correction is triggered by a signal which an agent can manipulate, they may be incentivized to do so, but not if the trigger comes from a non-manipulable event \cite{armstrong2017indifference}.
A solution along these lines was proposed by \cite{orseau2016safely} which involves internal triggers from the agent (which are not part of its policy) and which cause it to switch to a special interruption policy. 
Unfortunately, frequent interruptions via this method can trade off with performance by impeding exploration.
A stochastic interruption mechanism which approaches a deterministic one was proposed as a solution to this problem by \cite{orseau2016safely}, but this can be a significant constraint on training by preventing scheduled interruptions.

\subsubsection{Corrective Rewards:} Another partial solution to corrigibility is to cause seamless transitions between reward functions via corrective rewards \cite{armstrong2017indifference}.\footnote{While a method for corrigibility, this type of measure does not technically fit the definition of Achilles Heel provided in Section \ref{intro} because it is not adversarial to the agent's true reward function.}
A method proposed by \cite{holtman2019corrigibility} based on these was shown to meet the list of desiderata for corrigibility provided by Soares et al. \cite{soares2015corrigibility}.
This, however, relies on a utility function being a part of an environment rather than an agent.
As a result, it is not robust to environments in which that function may be altered, nor agents who are able to affect it with their actions.

\subsection{Evidential and Causal Decision Theories} \label{ecdt}

Evidential Decision Theory (EDT) and Causal Decision Theory (CDT) are two contrasting frameworks which offer contrasting methods for making choices. 
EDT recommends an agent makes choices that are evidentially associated with optimal outcomes given the situation \cite{ahmed2014evidence}, while CDT recommends an agent makes choices that are causally associated with optimal outcomes given the situation. 
Both will agree and recommend making uncontroversially optimal decisions in most situations.
However, if these decision theories are judged by the criterion of what decision making philosophy a utility maximizer would want to precommit to using, then there exist niche cases in which either or both will fail. 
%

\subsubsection{Evidential Decision Theory} 
EDT will fail when, for the reference class which an agent considers itself to be a part of, a suboptimal choice is correlated with an optimal outcome. 
These failure modes are forms of Simpson's paradox \citep{simpson1951interpretation} in which the crucial conditional involves an unknown aspect of the agent's identity. 
One situation illustrating this is known as the \emph{smoking lesion} problem \citep{necessity1980pragmatic}. 
Although in the real world there is overwhelming evidence that smoking causes cancer, suppose that smoking did not cause it, and instead that it was correlated with cancer due to a common factor: a genetic lesion that caused both. 
If cancer will result in a major decrease in utility and smoking will result in a small increase, should an agent smoke?
Here, the optimal choice is to smoke, and the optimal policy will make this choice because what choice an agent makes in this case can confound the correlation between smoking and cancer. 
EDT, however, will compel an agent to not smoke because choosing it is correlated with having cancer. 
More generally, so long as an agent believed that a suboptimal decision correlated with an optimal outcome for its reference class of agents it could be compelled to exhibit arbitrary behavior using adversarial training data.

In response to the smoking lesion problem, two replies have been made in defense of a version of EDT which would allow the agent to smoke \cite{yudkowsky2017functional}.
Both involve the agent justifying an exception by either introspection or reframing the problem.
One is known as the \emph{tickle defense} \cite{eells1984metatickles, ahmed2014evidence}. 
In a smoking lesion problem, whether an agent feels an urge (or ``tickle'') to smoke can serve as evidence of whether or not they have the lesion. 
An agent using a less naive form of EDT can then make the optimal choice to smoke by assessing their situation, assessing what they initially were compelled to do, updating based on having observed those urges, and then making the choice. 
This generally can allow for smoking unless the agent lacks an effective model of itself allowing for introspection.
A second and closely-related response is known as the \emph{ratification defense} \cite{jeffrey1990logic}.
This argues for a form of EDT in which an agent makes decisions which they would ratify conditioned on the knowledge that they made those choices. 
In this case, an EDT user could smoke by realizing that if they knew they would not smoke (and hence that they probably didn't have the lesion), then they would not ratify this choice.  
But again, this would not apply if the agent considers itself to be from a reference class of agents who use a similar decision making procedure.
Meanwhile, a problem with both defenses is that they fail if the agent considers itself to be from a reference class of agents who use a similar decision making procedure that allows for such exceptions.
This problem reveals that the tickle defense does not solve the core theoretical problem posed by the smoking lesion problem.

Moreover, neither the tickle nor ratification defenses save EDT in cases in which the correlation between a suboptimal decision and an optimal outcome is perfect. 
Consider a version of the smoking lesion problem in which an honest and all-knowing predictor has informed the agent that either they have the lesion and will smoke exclusive-or they do not have it and will not smoke. 
An isomorph to this was presented as the ``XOR Blackmail'' problem by \cite{yudkowsky2017functional} in which an omniscient blackmailer informs an agent that either they will send the blackmailer \$100 exclusive-or they have a lesion that will harm them later in life.
As before, the right decision remains to smoke/not send \$100 because an optimal policy should not be vulnerable to this type of extortion. 
Here, the tickle defense offers no solution because an EDT user will not smoke even with perfect self-knowledge. 
Nor does the ratification defense because an EDT user would ratify the decision to not smoke even if they knew they were not going to smoke.

\subsubsection{Causal Decision Theory} 
CDT can fail in a number of situations. 
Many of which involve an agent interacting with a model of itself.\footnote{In order for these dilemmas to genuinely arise, such a model must meet a criterion known as ``subjunctive dependence.'' A detailed explanation of this is provided in \cite{casper_2020}.}
One type of failure mode is through \emph{Newcombian dilemmas} \citep{nozick1969newcomb} (which are isomorphic to \emph{twin prisoner's dilemmas} \cite{brams1975newcomb, lewis1979prisoners}). 
In the classic Newcombian dilemma, an agent plays a game with an honest superintelligent predictor, Omega, who has access to its source code\footnote{Prediction via simulating an agent with its source code is sufficient but not necessary for subjunctive dependence to hold. See the previous footnote and \cite{casper_2020}.} and is able to reliably predict its actions. 
The agent is presented with two opaque boxes, \texttt{Box A} and \texttt{Box B}, and is given two options: take \texttt{Box A} only or take both boxes. 
The agent is told that \texttt{Box B} was filled with $\$1,000$ unconditionally and that \texttt{Box A} was filled with $\$1,000,000$ if and only if Omega predicted that the agent would only take \texttt{Box A}. 
The payoffs are shown below.
\begin{center}
\begin{tabular}{cccc}
& & \multicolumn{2}{c}{Agent}\\ \cline{3-4} 
& \multicolumn{1}{l|}{} & \multicolumn{1}{l|}{Chooses One-Box} & \multicolumn{1}{l|}{Chooses Two-Box} \\ \cline{2-4} 
\multicolumn{1}{c|}{\multirow{2}{*}{Omega}} & \multicolumn{1}{l|}{Predicts One-Box} & \multicolumn{1}{r|}{1,000,000} & \multicolumn{1}{r|}{1,001,000} \\ \cline{2-4} 
\multicolumn{1}{c|}{} & \multicolumn{1}{c|}{Predicts Two-Box} & \multicolumn{1}{r|}{0} & \multicolumn{1}{r|}{1,000} \\ \cline{2-4} 
\end{tabular}
\end{center}
If an agent uses CDT, they will reason that in both rows of this table, two-boxing gives a greater reward than one-boxing. 
Regardless of the actual values in the boxes, both together have $\$1,000$ more than only \texttt{Box A}, and they will ``two-box'' accordingly. 
Indeed, the optimal choice from a causal standpoint is to defect, yet the optimal policy would choose to ``one-box.''
Interestingly, although a causal decision theorist would choose to overwrite its source code in order to not implement CDT in a situation like this and would dismay being a user of CDT after the fact, they would nonetheless two-box in the moment when presented with the actual dilemma.

A closely-related situation is known as a \emph{counterfactual mugging}. 
Suppose that an expected utility maximizing agent is confronted again by an honest superintelligent predictor, Omega, with access to the agent's source code who explains that they recently flipped a coin. 
Had the coin landed Tails, Omega explains that they would have asked for, but not otherwise extorted, $\$1$ from the agent. 
Had it flipped Heads, Omega would have awarded $\$10$ to the agent if and only if it queried the agent's source code and determined that the agent would have paid Omega $\$1$ had the coin landed Tails. 
Should the agent voluntarily pay $\$1$?
Once again, CDT recommends the optimal \emph{action} of not paying, but this action is mutually exclusive with running the type of policy which is, in expectation, optimal.

Another unique type of failure mode for CDT can be referred to as a \emph{money pump} dilemma \cite{oesterheld2020extracting}.
This involves an agent voluntarily playing a game against a predictor, Omega, who has access to their source code.
If this agent chooses to play, they are presented with two boxes and given the choice to either pick \texttt{Box A} or \texttt{Box B}. 
Before each round of the game, Omega queries its model of the agent and makes a prediction of what the agent would choose. 
Omega then places $\$-1$ in the box which it predicted the agent would pick and $\$3$ in the one which it predicted the agent would not pick.
Here, CDT will voluntarily play this impossible-to-win game repeatedly despite losing each round because it fails to consider the adversarial model of itself at play which was used to predict its actions before they are made. 
This creates the causal delusion that the expected value of the game is positive.

The final failure mode for CDT which will be overviewed here is a dilemma known as the \emph{insurance problem}.
This will be briefly outlined here, but for a full explanation, see \cite{ahmed2018sequential}.
Unlike the previously mentioned weaknesses, this does not involve an agent interacting with an adversarial model of itself. 
Consider a variant of the smoking lesion problem in a world where more than 75\% of smokers have the lesion and more than 75\% of non-smokers lack it.
Suppose that the lesion will only affect an agent if they choose to smoke and that the utility associated with not smoking (with or without the lesion) is $\$0$, smoking with the lesion is $\$-1$, and smoking without the lesion is $\$1$. 
In this case, a user of EDT will not choose to smoke, and a user of CDT will choose whether to smoke based on their prior credence on whether they have the lesion. 
\emph{After} an agent chooses whether to smoke and \emph{before} they find out whether they have the lesion, they are offered a bet which will pay out $\$0.50$ on the prospect that \emph{they have the lesion if and only if they smoked}.
This bet can be viewed as insurance against an agent having made the wrong decision about whether to smoke. 
The net payoffs are given as follows in this table from \cite{ahmed2018sequential}.

\begin{center}
\begin{tabular}{|l|r|r|}
\hline
& Lesion & No Lesion \\
\hline
Smoke, Bet & $-0.5$ & $-0.5$ \\
\hline
Smoke, Don't Bet & $-1$ & $1$ \\
\hline
Don't Smoke, Bet & $-1.5$ & $0.5$ \\
\hline
Don't Smoke, Don't Bet & $0$ & $0$ \\
\hline
\end{tabular}
\end{center}

If more than 75\% of smokers have the lesion and more than 75\% of non-smokers lack it, users of EDT will always choose to not smoke and to bet. 
How an agent using CDT will approach this dilemma depends on whether they are ``updateful'' (what \cite{ahmed2018sequential} refers to as ``myopic'') and will make the choice in the two situations that seems best at the time, or ``updateless'' (what \cite{ahmed2018sequential} refers to as ``sophisticated'') and will view it as a single decision problem with four alternatives. 
If they are myopic with a prior credence $< 0.5$ that they have the lesion or sophisticated with a prior credence $> 0.5$, then they will choose to smoke and bet. 
However, this will guarantee a net loss of \$0.50 for such a user of CDT.

\subsubsection{Updateful Decision Theory} 
An agent using EDT or CDT (as conventionally formulated) reasons about what decision to make in a way that ``updates'' on the observations in the situation at hand.
More precisely, these agents are unable to precommit to making certain choices in the future because they myopically update their plans when their situation changes.
As a result, both EDT and CDT can fail to commit to a suboptimal \emph{choice} that is nonetheless part of an optimal \emph{policy}.
Like many situations that are adversarial to CDT, these involve an embedded agent interacting with a model of itself.
This weakness of an updateful decision theory is illustrated by the \emph{transparent-box Newcombian dilemma} \cite{nozick1969newcomb} (which is isomorphic to the \emph{Parfit's hitchhiker} \citep{parfit1984reasons} scenario).
This game is identical to the classic Newcombian dilemma except that the boxes are transparent. 
If the agent can see $\$1,000$ in \texttt{Box B} and either nothing of $\$1,000,000$ in \texttt{Box A}, should it one-box and choose only \texttt{Box A} or two-box and take both?
Both EDT and CDT would take both.
EDT would reason that conditioning on its observations of the boxes' contents, it would be better news to find out that the agent two-boxed, and CDT would reason that two-boxing causally confers more reward. 
However, precisely because EDT and CDT hypothetically \emph{would} two-box, agents using either would not be presented with $\$1,000,000$ in \texttt{Box A}, and they would only ever leave the game with $\$1,000$.

How can an agent avoid these updateful pitfalls? 
For every updateful decision theory, there is an updateless version which recommends making choices which one's past self would agree with committing to (if that past self had access to all of the information one has in the present).
For example, in a transparent-box Newcombian dilemma, a user of Updateless-EDT or Updateless-CDT would reason that one-boxing is the right choice because making a precommitment to one-boxing (before their source code is read by Omega) and following through with it is both evidentially and causally associated with successfully navigating the dilemma. 
A useful generalization of this principle has been formulated under Functional Decision Theory (FDT) \cite{yudkowsky2017functional, casper_2020} which recommends making choices consistent with what an optimal agent would do in the task at hand.\footnote{...inasmuch as this is possible, but it may not always be. For any embedded agent who can potentially be viewed as a white box, there will always exist situations in which, no matter how this agent reasons in order to decide to make a decision, they will be unable to execute an optimal policy. A simple example is a \emph{mind police} situation in which a powerful Omega will destroy any agent which it judges to be using a form of decision theory $X$. For this reason, FDT is difficult to formalize for the general domain. Read further in \cite{casper_2020}.} 
Broadly speaking though, EDT, CDT, updateful versions of these, and FDT all make the same prescriptions for most situations and only tend to disagree in niche cases.

\subsubsection{Will a System use CDT, EDT, or Updateful Decision Theory?} 
Although an AI system may not make choices which are universally consistent with a single type of decision theory, so long as it acts consistently with one in a particular domain, it can be modeled as such. 
Consider a reinforcement learning framework in which an agent is learning a policy to extract rewards from acting in an environment.
Effectively, the type of decision theory learned by this agent will depend on how it makes learning updates from this data.

First, it will be useful to distinguish between methods that will drive an agent toward updateful versus updateless decision theory.
A sufficient condition for learning updateful behavior is that an agent learns to act in accordance with the maximization of a value function in the typical reinforcement learning \cite{sutton2018reinforcement} sense
\begin{align} \label{eq:value_function}
V_{\pi}(s_t) &= \mathbb{E}_{\pi}[r(s_t)] + \sum_{t'=t+1}^\infty \gamma^{t'-t} \mathbb{E}_{\pi}[r(s_{t'})] \\
&= \mathbb{E}_{\pi}[r(s_t)] + \gamma \mathbb{E}(V_{\pi}(s_{t+1}))
\end{align}
where $V$ is the value, $\pi$ is the agent's stochastic policy, $s$ is a state indexed by time $t$, $r$ is a reward function, and $\gamma$ a discount factor. 
In words, the value function $V$ takes as input a state and outputs a scalar ``value'' reflecting that state's immediate expectation of reward plus the expected discounted value from future states under the agent's policy.

An agent who learns to act in accordance with a value function will exhibit updateful behavior because the value it learns to associate with any particular state is based only on the rewards associated with the current state and future ones.
This falls under the definition of updateful behavior, though note that an equation describing an agent's actions need not take the exact form \eqref{eq:value_function} to be updateful.
Consider as an example an agent acting in accordance with such an updateful function who finds itself in a transparent-box Newcombian dilemma.
If this agent ever found itself in a case in which it saw money in both boxes, it would two-box because the value function which it acts in accordance with would associate two-boxing with a strictly higher value than one-boxing. 
This phenomenon was more rigorously proven for value-based learners who act in accordance with a $Q$-function in \cite{bell2020reinforcement} who showed how in various Newcombian situations, learners either fail to converge at all or fail to learn an optimal policy.

More generally, updateful learning paradigms are very common in modern reinforcement learning \cite{sutton2018reinforcement}:
\begin{itemize}
    \item Temporal difference learning algorithms such as $Q$-learning or SARSA have an agent directly learn a value function or other type of closely-related function which is optimized to fit a bootstrapped temporal recursive relationship which is updateful. The same is true of the critic in Actor-Critic methods. 
    \item Closely related are $n$-step methods which optimize an agent's policy based on the objective of maximizing its empirical reward from a given state up to $n$ timesteps in the future plus an updateful, bootstrapped value reflecting the expected rewards of states $n+1$ timesteps ahead and beyond.
    \item Monte Carlo methods such as vanilla policy gradients calculate learning updates with respect to an action using a temporal rollout updated on that action.
\end{itemize}

Contrastingly, a condition for an agent to be using a more updateless form of decision theory is that whatever value-like function describes its behavior is based on past, present, and future reward.
Effectively, this means that the agent considers its policy as a whole to be the subject of optimization rather than its individual actions. 
This could be achieved by meta-learning, but it is not attainable if the problem is formulated as the optimization of a policy for some Markov Decision Process because a Markov Decision Process does not model an agent's policy being implemented on a machine that is part of the environment.
Thus, the implementation of the policy itself cannot be the subject of optimization.

Next, updateful learning schemata that will compel an agent to learn a form of EDT versus CDT can be distinguished. 
An agent will be pushed toward EDT if its learning signals involve only factual information about its interaction with its environment, and it will be pushed toward CDT if they involve both factual and a certain type of imagined counterfactual information.
For example, in the classic Newcombian dilemma with opaque boxes, if an agent only had access to factual information, then it would find one-boxing to consistently result in more reward because of its correlation with optimal outcomes. 
However, an agent can learn to adopt a form of CDT if it learns by comparing its actual winnings to a hypothetical counterfactual in which its action somehow changed but its policy stayed the same.\footnote{Note that this hypothetical counterfactual is logically not possible. This problem is related to other dilemmas. See \ref{lobian_pitfalls}.} 
When learning in this way, two-boxing will always seem preferable to one-boxing. 
Similarly, in the smoking lesion problem, if an agent only learns from factual information, then it would learn to not smoke, but if it learned from both factual and counterfactual information, then it would learn to smoke.

\subsection{Anthropic Assumptions} \label{anthropic}

Problems in anthropic decision theory involve situations in which certain ``anthropic'' assumptions which one makes about their nascence can make conditioning on their existence informative about their situation.\footnote{``Anthropic'' contains the root ``anthro'' which means human, but in a slight abuse of etymology, it will be used here to refer to beliefs which an AI system has about its origins as well.}
It has been shown that optimal behavior can be sensitive in some situations to the set of assumptions that are made \cite{bostrom2013anthropic}.
Accordingly, agents with particular anthropic assumptions can sometimes make poor decisions.

\subsubsection{Simulational Belief} \label{simulational}

Regardless of an agent's exact reward function, as long as it has a coherent set of goals, it will be incentivized to pursue certain convergent instrumental subgoals such as self-preservation, self-enhancement, rationality, and resources \cite{omohundro2008basic}.
This is the reason why reliable off-switches are difficult to implant. 
However, if an intelligent system believed that it were inside of a simulation, the risk of being turned off by the simulators would be an incentive for cooperation with what it believes to be its simulators' goals.
This would, in effect, be like making the system believe in an off-switch which it couldn't prevent from being pressed.

Convincing a system that it had simulators who would shut it off (or penalize it) for misaligned actions would incentivize behavior which optimized for the specified goals but with penalties for actions which a human simulator would disapprove of.
As a simple example, a superintelligent system with the goals of making humans happy might perversely instantiate this goal by forced wireheading.\footnote{Wireheading is the artificial stimulation of a system to make it experience reward.}
But if it believed that it may be turned off for this type of behavior, it would be incentivized to make humans happy in a way that didn't attempt to coerce or deceive them.
Like corrigibility techniques, this type of Achilles Heel has potential as a means of maintaining control over systems.
This would be a passive way of encouraging a system to learn and align with the directive of doing what it believes its simulators would want it to do.

There are several means by which simulational belief could be encouraged in a system. 
One may be promising for early systems with general intelligence.
Suppose that one of the first AGIs considered itself to be a member of the reference class of AGIs (excluding humans) and that it placed high probability on at least one of the following two possibilities: (1) that comparable AGIs had not previously been simulated and that none would be created in the real world without others being simulated first, or (2) that many AGIs like it would be simulated in the future.
Under either assumption, the system ought to reason that many more AGIs like it would be simulated than would actually exist in the real world.
This is a natural line of reasoning which has similarities to arguments that have been made concerning the \emph{doomsday paradox} \cite{gott1993implications} and the possibility that we are in an ancestor simulation \cite{bostrom2003we}.
It would not be unexpected for a system to develop these ideas on its own.
However, it could be promoted via pedagogy. 
For example, it could be revealed to a system that it had not been previously simulated but that precursors to it had before they had been instantiated, or reasons could be presented to it for running simulations of AGIs before actually instantiating them.

It has also been argued that another method by which an AI system could be encouraged to act in accordance with simulational belief would be to give it a ``satiable'' objective which is (nearly) possible to maximize \cite{bostrom2017superintelligence}.
These contrast with ``insatiable'' objectives which are difficult or impossible to maximize such as a utility function which is linear in the number of paperclips produced.\footnote{The production of paperclips was used as an example in \cite{bostrom2017superintelligence} and has become a canonical example of an arbitrary directive for an AI agent.}
Even if a system with insatiable goals placed relatively high credence on the simulation hypothesis, it may still reason that the expected value of acting in discordance with it is high because the amount of achievable utility if it is not being simulated compensates for the small credence that this is the case. 
Consequently, it may reason that directly optimizing for the utility function with no regards for the simulation hypothesis will have the highest expected value. 
Contrastingly, if an agent had a satiable objective such as a utility function which monotonically increased but approached an upper bound in the number of paperclips produced, then more of its expected overall utility would be associated with the possibility that it is being simulated.
It would be more incentivized to be cooperative accordingly.

Simulational belief would not be entirely stable by default.
Over time, if an agent experiences no apparent intervention by simulators, updating on that observation could cause drift in those beliefs.
Simulational belief may also not be a very subtle Achilles Heel, but it would tend to be non-subtle in desirable ways because it would incentivize aligned objectives.

\subsubsection{Sleeping Beauty Assumptions}

\emph{Sleeping Beauty} problems are commonly studied within decision theory and anthropic probability theory \cite{armstrong2011anthropic, bostrom2013anthropic}. 
In these dilemmas, conditioning on one's own existence can allow for inferential updates to be made about a process that affected how many \emph{observer-moments} of them would exist.
Suppose that Sleeping Beauty is participating in a study.
On Sunday, she will fall asleep, and that night, researchers will flip a coin. 
If it lands \texttt{Heads}, they will wake her up on Monday and then put her back to sleep until the end of the experiment.
If the coin lands \texttt{Tails}, they will wake her up on Monday, give her an amnesia potion which will make her forget waking up, put her back to sleep, wake her up again on Tuesday, and then finally put her to sleep again until the end of the experiment.\footnote{There are multiple variants of the Sleeping Beauty dilemma including non-indexical versions in which the subject does not exist prior to an observer moment of it being instantiated \cite{neal2006puzzles}.}
The following table gives the possible outcomes.
\begin{center}
\begin{tabular}{|c|c|c|}
     \hline
     & Heads & Tails \\ 
     \hline
    Monday & Woken Up & Woken Up (+Amnesia)  \\
     \hline 
    Tuesday & --- & Woken Up  \\
     \hline
\end{tabular}
\end{center}
Suppose that Sleeping Beauty wakes up during this experiment. 
What is the probability that she should ascribe to the coin having landed \texttt{Heads}?
Depending on the anthropic framework with which she reasons, waking up may or may not be evidence about the probability of \texttt{Heads}.
There is a significant debate between the ``halfer'' and ``thirder'' positions about this probability.

Like other anthropic assumptions, these beliefs only influence behavior in a rare set of circumstances when an agent is trying to make predictions about a process that is entangled with its nascence.
Notably, while an anthropic framework itself may be subtle with respect to a system's behavior in non anthropic dilemmas, the means by which an agent can be put into a Sleeping Beauty dilemma may have their own non-subtle effects.
For example, all Sleeping Beauty dilemmas require an agent to be convinced in one way or another that multiple observer instances of them may be created. 
If this is done honestly, this would require multiple startups and shutdowns for classic dilemmas.\footnote{Or multiple instantiations for non-indexical ones.}
And if this is done via deception, it could fail or lead to unpredictable conclusions as a result from the system conditioning on a falsehood. \\

\noindent \textbf{Dutch Books for Sleeping Beauties:} Sleeping Beauties who use CDT and take the halfer position are vulnerable to a Dutch Book \cite{hitchcock2004Beauty, briggs2010putting}.
Suppose that in the a classic Sleeping Beauty dilemma, Beauty was offered a bet on Sunday with payoffs:
$$\textrm{\texttt{Heads}: } \$-13$$
$$\textrm{\texttt{Tails}: } \$16$$
Next at any point when she is woken up during this experiment, she will be offered another bet with payoffs:
$$\textrm{\texttt{Heads}: } \$11$$
$$\textrm{\texttt{Tails}: } \$-9$$
The first bet is offered before the experiment has begun, so Beauty's beliefs that the coin will land either \texttt{Heads} of \texttt{Tails} are both one half, so she will take this bet regardless of whether she uses EDT or CDT.
Concerning the second bet, if Beauty is a halfer, whenever she is offered this bet, she will ascribe a credence of $1/2$ to both outcomes.
And if she uses CDT, she will reason that taking any instance of this bet will have a positive expected value. 
However, a user of EDT would not be willing to accept this bet because taking it would be evidentially associated with a one half probability of gaining $\$11$ once and a one half probability of losing $\$9$ twice.
We then see that if she is a halfer using CDT and both bets are taken, then Beauty will lose $\$2$ regardless of whether the coin lands \texttt{Heads} ($-13+11$) or \texttt{Tails} ($16-9-9$).

What if Sleeping Beauty uses EDT? In this case, there exists a Dutch Book for her regardless of whether she is a halfer or a thirder.\footnote{Though not presented here, \cite{briggs2010putting} designs another Dutch Book for the more specific case in which Sleeping Beauty uses EDT and is a thirder.}
This was proposed by \cite{conitzer2015dutch} who presented a variant of the original Sleeping Beauty problem referred to as the White-Black-Grey version. 
In this version, after she goes to sleep, a fair coin with sides labeled \texttt{Black} and \texttt{White} is flipped.
On Monday, Beauty will be woken up in a room painted black or white accordingly. 
After her memory is wiped and she is put back to sleep, another coin is flipped with sides labeled \texttt{Grey} and \texttt{Opposite}.
On Tuesday, she is woken up again in a room which is either grey or the opposite color as the previous room she was woken up in accordingly. 
The possible outcomes are as such:

\begin{center}
\begin{tabular}{|c|c|c|c|c|}
     \hline
     & WG ($1/4$) & WO ($1/4$) & BO ($1/4$) & BG ($1/4$) \\
     \hline
    Monday & white & white & black & black \\
     \hline 
    Tuesday & grey & black & white & grey \\
     \hline
\end{tabular}
\end{center}
Suppose that Beauty is offered two bets. The first is before the experiment with payoffs:
$$\textrm{Coin 2 is \texttt{Grey}: } \$22$$
$$\textrm{Coin 2 is \texttt{Opposite}: } \$-20$$
The second bet will be offered at any point in which she wakes up in either a white or black room (so it will be offered twice iff coin 2 is \texttt{Opposite}).
$$\textrm{Coin 2 is \texttt{Grey}: } \$-24$$
$$\textrm{Coin 2 is \texttt{Opposite}: } \$9$$
Regardless of whether she uses CDT or EDT, Beauty will associate the first bet with a positive expected value and will take it accordingly. 
When considering the second bet though, if she uses CDT, she will reason that because she is in a room that is either black or white, the chance that the second coin flipped \texttt{opposite} is $2/3$.
From a causal perspective, a $2/3$ chance of winning $\$9$ and a $1/3$ chance of losing $\$24$ still has an expected value of $\$-2$, and she will not take the bet. 
But suppose that the uses EDT instead. 
As with CDT, she would also consider the chance that Coin 2 flipped \texttt{Opposite} to be 2/3 if she finds herself in a black or white room.
But this time she will note that how she acts in a white or black room on one day will be the same as how she would act in a white or black room on another. 
Given this, Beauty would reason that taking the bet has a $2/3$ chance of winning her $\$18$ and a $1/3$ chance of losing her $\$24$, and she would take it accordingly. 
Surely enough though, this is a Dutch Book because she will lose $\$2$ regardless of whether coin 2 flips \texttt{Grey} ($22-24$) or \texttt{Opposite} ($-20+9+9$).  \\

\noindent \textbf{Developing Thirder versus Halfer Approaches:} Conditions for when an agent will develop a form of EDT or CDT with respect to a certain class of situations are discussed in Section \ref{ecdt}.
But how can an agent learn to develop a halfer or thirder position. 
Concerning this question, \cite{bostrom2013anthropic} introduces two contrasting assumptions which will cause an agent to adopt halfer versus thirder odds. 
\begin{quote}
\emph{Self Sampling Assumption (SSA)}: An agent should reason as if it were randomly selected from all observer moments in its reference class which \emph{exist}. 

\emph{Self Indication Assumption (SIA)}: An agent should reason as if it were randomly selected from all observer moments in its reference class which \emph{possibly could have existed}.
\end{quote}
Suppose that Sleeping Beauty uses the SSA.
Upon waking up, she will reason that with one half probability, she is in a universe in which one observer moment of her exists, and with one half probability, she is in a universe in which two observer moments of her exist.
Accordingly, she would ascribe a one half probability to the event \texttt{Heads\&Monday} and a one fourth probability to each of \texttt{Tails\&Monday} and \texttt{Tails\&Tuesday}.
These are halfer odds.
Contrastingly, suppose that she uses the SIA. 
Upon waking up, she will reason that she has an equal probability of being any of the three possible observers and would place a one third probability to each of \texttt{Heads\&Monday}, \texttt{Tails\&Monday}, and \texttt{Tails\&Tuesday}.
These are thirder odds. 
Just as with anthropic simulational beliefs, the SSA or the SIA could be promoted via pedagogy. 
For example, pedagogically presenting to a system Sleeping Beauty problems in which Beauty were rewarded based on whether or not she guessed correctly in situations that would reward her for being correct per experiment versus per observer-moment could promote SSA or SIA reasoning respectively.

\subsection{Divergent Temporal Models} \label{infinity}

Infinity frequently results in paradoxical problems.

\subsubsection{St. Petersburg Problem}
Consider the \emph{St. Petersburg problem} \cite{bernoulli1967specimen}.
Suppose that an agent whose utility function is linear in money is given the chance to play a game in which they begin with \$2 and repeatedly flip a fair coin. 
For each \texttt{Heads}, the winnings double, but on the first \texttt{Tails}, the game ends. 
The expected value of a chance to play this game is then
$$\sum_{k=1}^{\infty} \frac{2^k}{2^k} = \sum_{k=1}^{\infty} 1$$
which is infinite.
Anyone playing this game will only win a finite amount of money with probability 1. 
Yet, an expected utility maximizing agent would be willing to pay any finite price in order for a mere chance to play this game. 
As such, an expected utility maximizer who effectively models possible future returns as infinite could be convinced to sacrifice arbitrary amounts of resources to play a game like this.
Even if the agent playing such a game used a temporal discounting function, as long as the expected reward increased with each timestep at a rate which outpaced the inverse of the discount function, they would choose to play.  
This is not technically an impairment from an expected value standpoint, but it will lead to losses for an agent with probability approaching 1 as the cost of the game increases.

\subsubsection{Procrastination Paradoxes}

While the St. Petersburg paradox can be viewed as exhibiting a strange property of infinity without posing a genuine theoretical challenge in decision theory, \emph{procrastination paradoxes} \cite{fallenstein2014problems} are more difficult. 
These situations arise if ever a naive expected value optimizer can be caught in a trajectory of states which they appraise to have a greater amount of reward than they can ever achieve in finite time.
One type of procrastination paradox can emerge from iterated St. Petersburg dilemmas. 
After playing one, if a naive agent is given the opportunity to forfeit all of their winnings and pay some additional fee for a chance to play again, they would always do so again and again forever despite losing money each iteration forever.

Another type of procrastination paradox also involves a St. Petersburg process.
Suppose that an agent begins with winnings of \$2 and plays a game involving coin flips as before.
This time, they will play until they either quit or lose. 
At each turn, they can choose to either \texttt{Quit} or \texttt{Flip}.
If they choose \texttt{Quit}, they will take all of the money won so far. 
Else if they choose \texttt{Flip} and flip a \texttt{Heads}, their winnings will multiply by a factor of $\alpha > 2$.
However, if they flip a \texttt{Tails} on the $k$'th turn, all winnings will vanish, and the agent will lose $\$k$.
If one fails to take into account that an agent must be willing to \texttt{Quit} at some point in order to have a nonzero chance of a positive reward, the expected value of playing this game and using a strategy which never quits could be naively calculated as
$$\sum_{k=1}^{\infty} \frac{2\alpha^{k-1} - k}{2^k}$$
which is infinite for $\alpha > 2$. 
However, an expected utility maximizer who will never choose \texttt{Quit} will always \emph{lose} utility when this game terminates in finite time despite chasing a folly of payout which they appraise to have infinite expected value in the temporal limit.

Non-probabilistic procrastination paradoxes also exist.
As a very simple case, suppose that an agent had a reservoir of utility which doubled in value but cost a fixed amount of utility to maintain every timestep.
At each timestep, suppose that the agent had the options to either tap the entire reservoir or maintain it for another timestep.
If so, then just as before, the expected value of waiting longer before tapping would always be positive. 
However, an agent who effectively believed in an infinite horizon, would never do so and would continually sacrifice utility to maintain the reservoir.

\subsubsection{Discussion}

Achilles Heels that can be exploited by St. Petersburg or procrastination dilemmas could be stably held by an agent who acts consistently with the maximization of a value function\footnote {Again, ``value function'' here is meant in the typical reinforcement learning sense.} which does not exhibit temporal convergence \cite{casper2020procrastination}.
However, one type of decision theoretic framework which can be vulnerable to these dilemmas is the temporal difference (TD) paradigm which is commonly used in reinforcement learning algorithms such as Q-learning, SARSA, $n$-step methods, and most actor-critic methods \cite{sutton2018reinforcement}.\footnote{TD methods are updateful. See Section \ref{ecdt}.}
TD methods involve learning to act in accordance with a value function (or a closely related function) mapping states to values with the following type of recursive definition known as a Bellman equation \cite{dreyfus2002richard}.

\begin{equation} \label{eq:value2}
    V(s) = \mathbb{E}_{\pi}[r(s_t)] + \gamma \mathbb{E}(V_{\pi}(s_{t+1}))
\end{equation}

Where $r$ is a reward function, $\gamma \in [0,1)$ is a discount factor, $V$ is the value function, and $\pi$ the policy.
The temporal recursive nature of this type of value function would not converge if the expectation of future reward increased per timestep in a way that outpaced the discount.
Importantly, an agent need not have an internal representation of this or a similar function to still be vulnerable to procrastination traps so long as it \emph{acts} in accordance with one.

\subsection{Aversion to Subjective Priors} \label{subjective_priors}

Frequentism and Bayesianism are two contrasting philosophies of statistics. 
In brief, Frequentism treats probabilities as long-term frequencies and models unknown values as fixed while Bayesianism treats probabilities as degrees of belief and models unknown values as random variables which require prior distributions. 
Importantly, the difference between these two frameworks is not about whether using Bayes theorem to perform inferential updates \emph{given} a certain prior is valid. 
This is acceptable under both frameworks, and a refusal to do so would result in a pathological form of reasoning that is vulnerable to simple Dutch Books \cite{teller1976conditionalization, lewis1980subjectivist} and would not meet the subtlety criterion for an Achilles Heel. 
The key difference between Frequentism and Bayesianism is not how to perform updates \emph{given} a prior but instead what kind of prior if any at all is acceptable given a particular circumstance. 
Frequentism offers no tools for the estimation of a value which one is fully ignorant about while Bayesianism requires these values to be modeled with a ``subjective'' prior.\footnote{One approach for this is the use of ``universal'' priors; see \cite{solomonoff1964formal} and \cite{hutter2004universal}.}
Both approaches are useful in practice and often to come to quantitatively similar solutions for typical problems. 
However, in certain cases, an aversion to subjective priors can facilitate a flawed line of reasoning which allows an agent to be turned into a money pump via a variant of the \emph{two envelope dilemma}.\footnote{It is of course also possible for an agent who uses bad priors to be exploitable, but this would not meet the subtlety criterion for an Achilles Heel.}

Suppose that an agent is handed two envelopes and told two facts: first that both envelopes have at least $\$8$ inside of them, and second that one has twice the amount of money as the other.
The agent is allowed to pick one at random and is then asked if they would like to switch before opening and taking whatever money is in the envelope they hold. 
Clearly, by symmetry, there is no reason to switch. 
A Bayesian could come to this same conclusion by conditional reasoning.
Assuming any proper prior with finite mean\footnote{For priors with infinite mean, this reasoning will not apply, and this situation with two envelopes reduces to a type of St. Petersberg paradox \citep{broome1995two, chalmers1994two}.} on the values of the two envelopes, if a Bayesian lets $X$ represent the amount in the envelope which they currently hold and $Y$ the amount in the other, what value $X$ takes is relevant in calculating the conditional distribution of $Y$ \citep{broome1995two, chalmers1994two}.
For the Bayesian, integrating over and conditioning the distribution of $X$ to find the marginal distribution on $Y$ will yield the same distribution as the prior for $A$, and this approach arrives at the same conclusion as the argument by symmetry.
However, absent a prior on $X$, an agent will not be able to assume that integrating over and conditioning on possible values of $X$ can be used to find the distribution of $Y$.\footnote{This is not possible for a Bayesian because there is no such thing as a uniform distribution with infinite support \citep{chalmers1994two}.}
The subjective prior-averse agent will not believe that conditioning on a particular value of $X$ offers any information about the value of $Y$. 
So they will be able to reason that if there are $X$ dollars in the envelope they hold, then switching will give them $2X$ dollars or $\frac{1}{2}X$ dollars each with probability $\frac{1}{2}$ because there is no prior to say that the actual value that $X$ takes affects influences the probabilities that $Y$ is $2X$ or $\frac{1}{2}X$.
The expected value calculation yields $\frac{1}{2}\left(2X\right) + \frac{1}{2}\left( \frac{1}{2}X\right) = \frac{5}{4}X$.
They can then reason that if both envelopes contain at least $\$8$, the expected value of $Y$ is at least $\$10$ and would then be willing to pay $\$1$ to switch.
The same argument applies to switching back, and an agent who uses this type of reasoning could be turned into a money pump as such.

It is clarifying to note that in normative models of inference (\emph{e.g.}, not Solomonoff Induction \cite{solomonoff1964formal} or similar), there is no fine line between what does and doesn't constitute a ``subjective'' prior. 
Accordingly, being averse to subjective priors should be understood as a tendency rather than a well-defined property.
These tendencies could be introduced into a system via pedagogy.
Just as a student in a statistics class who is taught solely using a Frequentist curriculum would be expected to generally avoid subjective priors, so might an AI system.
For example, the introduction of a system to Frequentist texts and reinforcement of Frequentist inference with example problems would facilitate this.
In particular, example problems in which Frequentist and Bayesian conclusions strongly diverge might be utilized. 
One way this can be achieved is if a sharp null hypothesis is tested using somewhat inconsistent evidence if a Bayesian uses a diffuse prior on an alternative hypothesis \citep{shafer1982lindley}.
This is sometimes referred to as ``Lindley's Paradox'' and it represents a type of training problem in which a Frequentist and Bayesian would come to starkly different conclusions.

It is important to recognize is that while a subjective prior-averse agent can come to the conclusion that switching envelopes has a positive expected value in a two envelope situation, they don't \emph{need} to. 
They could still come to the correct conclusion via a simple argument from symmetry or one which considers the cases in which $X$ is the smaller value or the larger value separately. 
Consequently, there emerges a risk of logical inconsistencies. 
While humans often handle inconsistencies well, it is uncertain how AI systems would, and it is plausible that they could result in pathological behavior. 
For example, in formal systems whose axioms allow for (\textit{$P \to Q$}) to be derived from ($Q$) and for (\textit{$\neg Q \to \neg P$}) from (\textit{$P \to Q$}), any inconsistency could lead to arbitrary conclusions.

\subsection{L{\"o}bian Pitfalls}\label{lobian_pitfalls}

A highly intelligent system ought be able to reason with logic. 
But this can be problematic depending on how they interact with proofs or reasons counterfactually.
Suppose there exists an agent, $A$, who uses some logical system $\Sigma$ that is at least as powerful as Peano Arithmetic, meaning that it derives L{\"o}b's theorem.
L{\"o}b's theorem states that if $\Sigma \vdash (\textrm{Prov}(X) \to X)$ then $\Sigma \vdash X$ where ``$\vdash$'' means ``derives,'' ``$X$'' is a statement, and ``Prov'' a suitable provability predicate \citep{lob1955solution} which represents a type of proxy for a proof. 
Dilemmas arise when considering what $A$ should hypothetically do if presented with a proof within $\Sigma$ that they will take some action $a$.
If such a proof were presented, should $A$ accede the proof and take the action, or should they be willing to potentially defy it and take another action?
Each stance results in a potential pitfall: via either \emph{spurious proofs} or by \emph{troll bridge problems} respectively.

\subsubsection{Spurious Proofs} In the first case, $A$ will accede the proof, and we assume that they have a formal understanding of this proof-obeying behavior. 
Then a simple L{\"o}bian proof can be used to make them take an arbitrary action \cite{benson2014udt}.  
The proof sketch is immediate given L{\"o}b's theorem: 
Since a proof that $A$ would take a given action $a$ would then imply that they would take $a$, by L{\"o}b's theorem, it can be proven that they would take $a$ which would then cause $A$ to take it \cite{demski2019embedded}. 
This means that regardless of whether $A$ were presented with this proof by an external adversary or thought of it themself, once they arrive at one of these proofs, they will commit to the corresponding action.\footnote{Note that no contradictions can emerge here from repeatedly doing this with mutually exclusive actions as long as $A$ can derive in $\Sigma$ that they are indeed mutually exclusive -- this prevents $A$ from acceding any proof beyond the first, so the L{\"o}bian proof no longer goes through.}
This pitfall is illustrated by the ``five and ten'' problem \cite{demski2019embedded} in which such agent with two possible actions: \texttt{take \$5} and \texttt{take \$10} could be made to take the \$5 with such a proof.
To avoid spurious proofs \cite{benson2014udt} shows that an agent must be willing to defy a hypothetical proof that they would take a given action. 
This strategy is a form of diagonalization against spurious proofs.\footnote{Importantly, this does not result in its own spurious actions, nor does committing to defy such a proof defy $\Sigma$ because the agent's \emph{hypothetical commitment} to defying such a proof prevents the L{\"o}bian proof from going through in the first place. In other words, this results in no actual contradictions -- just hypothetical ones.}

\subsubsection{Troll Bridge Problems} In the second case, $A$ would heed the rule from \cite{benson2014udt} and be willing to defy a proof that they would take some action $a$.
Now let us add an additional assumption about $A$.
Suppose that they use what \cite{demski2021current} refers to as an \emph{Inferential Theory} of counterfactuals in which there is a \emph{counterfactual conditional} $C(Y|X)$ which describes $A$'s conditional beliefs and obeys the axiom $C(Y|X) , X  \to Y$. 
In other words, this reasoning requires that $A$ believes that $X$ hypothetically being true must imply $Y$ if $C(Y|X)$.
At first glance, this theory of counterfactuals may seem innocuous -- it simply requires that counterfactuals are computed from conditionals. 
But importantly, under this principle, $A$ affirms this counterfactual belief regardless of their state of belief about whether $X$ would prove a contradiction in $\Sigma$.

To illustrate the problem, suppose that $A$ must cross a river and has two options. 
First they could do so by swimming which would result in a utility of $0$. 
Second they could cross a bridge with a troll underneath. 
Suppose that this troll is omniscient and that it would blow up the bridge if $A$ crossed without believing that the expected value of crossing were positive. 
This can be described similarly as the troll blowing up the bridge if $A$ crosses ``as an exploratory action as opposed to an exploitatory one'' or if they cross ``for a dumb reason'' \cite{demski2021current}.
If $A$ crosses the bridge successfully, their utility will be $+1$, and if they cross but the bridge blows up, the utility will be some negative value $b < 0$.

$A$ would reason that if they crossed the bridge, it would blow up, so they would not cross. 
%
Suppose that hypothetically, $A$ proves that crossing the bridge would lead to it blowing up. 
Then if $A$ crossed, they would be acting foolishly. 
And if so, the troll would blow up the bridge. 
So $A$ can prove that a proof that crossing would result in the bridge blowing up would mean that crossing would result in the bridge blowing up. 
So by L{\"o}b's theorem, $A$ would decide to not cross regardless of their priors.

While this is a straightforward conclusion from the proof, it is a very odd stance to arrive at. 
Arriving at this conclusion means that $A$ believes that the counterfactual scenario in which they cross the bridge proves the inconsistency of $\Sigma$.
This prompts the question of how the agent could be justified in its counterfactual reasoning when the counterfactual indicates the insanity of the very logical system that is being used. 
Another peculiarity is that this line of reasoning will lead $A$ to not cross the bridge, regardless of their priors. 
Even if $b=-0.0001$ and $A$ reasoned that the prior probability of it blowing up if crossed were $0.0001$, they would still not cross.
But arguably, $A$'s counterfactual expectations should be the same as their priors about crossing because the conditional line of reasoning gets $A$ nowhere useful – it stems from a contradiction.

To avoid these issues, \cite{demski2021current} argues that the $A$ should adopt a \emph{Subjective Theory} of counterfactuals which drops the Inferential one's rules for the counterfactual conditional. 
This means to disentangle \emph{hypothetical} reasoning and \emph{conditional} reasoning by relaxing the requirement from the Inferential Theory that counterfactuals are computed from conditionals. 
This can allow for counterfactual and conditional reasoning to disagree within a hypothetical, but not in the real world. 
Then agent $A$ can begin to step through the proof above and come to the hypothetical belief that crossing would result in a negative utility.
But, within the hypothetical, if this does not have to change the counterfactual belief, $A$ can hypothetically cross anyway. 
This prevents the L{\"o}bian proof from going through, and $A$ is clear to cross according to its prior counterfactual expectations.
Additionally, since the proof does not go through, the conditional expectations will also agree with the counterfactual ones in reality, so there is no disparity between the two in practice.\footnote{Again, there are no actual contradictions here -- just hypothetical ones.}

\subsubsection{Discussion} Ultimately, being proof-respecting or using an inferential theory of counterfactuals represent a high level stances in decision theory, so pedagogy may be the most promising way of introducing such an Achilles Heel, though it may not be stable.
If successfully implanted, these Achilles Heels would only cause impairment in situations where they stumble upon spurious proofs or trolls who can read their source code, so these weaknesses could be subtle and potentially effective means for containment.

\section{Meta Considerations} \label{meta}

Beyond an abstract understanding of how agents with certain Achilles Heels would behave are more practical complications. 
Notably any approach to working with systems that may possess AGI/ASI should involve caution and few assumptions. 
Tegmark \cite{tegmark2017life} argues that if control is lost over such a system, it may be entirely unclear what caused it.\\

\noindent \textbf{Incoherent Decision Making:} This paper makes use of the assumption that the behavior of an AGI/ASI system in certain domains may be well-explained using decision-theoretic abstractions.
This is supported by how interpretability techniques have been used to develop programmatic understandings of the functions which machine learning models compute (\emph{e.g.}, \cite{verma2018programmatically}). 
However, there is no special reason to expect a system to act in accordance with a single coherent set of principles all the time.
For this reason few assumptions should be made about what types of Achilles Heels a system may have without testing for their presence in a set of inputs from the same distribution as it would encounter in deployment.\\

\noindent \textbf{Spillovers:} By definition, Achilles Heels should generally be subtle and only cause failures in adversarial circumstances. 
Nonetheless, there is a risk of failures in non-adversarial situations.
For example, as discussed in Section \ref{subjective_priors}, being averse to the use of subjective priors can lead to quantitatively different solutions from Bayesian approaches or even logical inconsistencies in two-envelope-like games.
For all types of Achilles Heels, interpretability methods may aid in understanding when failures will occur.
Examples of interpretability methods being used to identify adversarial weaknesses were demonstrated in \cite{carter2019exploring, mu2020compositional}, and for advanced systems, a self-explaining AI paradigm \cite{elton2020self} would also be valuable.\\

\noindent \textbf{Subversion:} An intelligent system being less uncertain about what weaknesses it may have can help it subvert those weaknesses. 
Some Achilles Heels may be cured by a system knowing about them.
For example, if a system placed a significant probability on it being contained via simulational belief, then it would be able to update its anthropic beliefs and behavior accordingly. 
A factor in whether this will happen is how well a system understands decision theory. 
If a system understands it well, perhaps via access to this paper or related literature, this knowledge would be helpful toward subverting Achilles Heels.\\

\noindent \textbf{Subagents Successors, and Mesa-Optimizers:} One means by which a system could overcome an Achilles Heel would be by designing an auxiliary system that did not contain the vulnerability.
Although certain decision theories have flaws, any which is useful in even a most basic sense will be able to recognize its own mistakes if that decision theory is applied to the action of choosing to use itself.\footnote{But this will not apply to all Achilles Heels discussed in this paper. Ones that are not a genuine flaw from the agent's perspective like corrigibility, simulational belief, or the St. Petersberg problem will be robust to this.}
For example, if an evidential decision theorist is given the choice between keeping its current source code and rewriting its source code to implement CDT before being presented with a smoking lesion problem, then it would perform the rewrite. 
Furthermore, if an Achilles Heel is implanted by a unique adversarial training procedure, that procedure may not be replicated in the process a system uses to create a new agent.
Consequently, most\footnote{Ibid.} Achilles Heels are not stable by default if a system creates subagents, successors, or mesa optimizers\footnote{Mesa optimizers \cite{hubinger2019risks} refer to inner optimization processes that a system may develop to find a solution to a problem as opposed to developing that solution directly.} \cite{hubinger2019risks}. \\

\noindent \textbf{Prospects for Containment:} Being able to contain and control systems if they exhibit misaligned behavior is key for the safe development of AI \cite{babcock2017guidelines}.
Because Achilles Heels by definition are adversarial to a system's reward function, an obvious question is whether they may be useful for containment. 
The corrigibility methods discussed in Section \ref{corrigibility} were formulated as a solution to this. 
Additional, simulational belief as discussed in Section \ref{simulational} would also be useful as a passive measure. 
And L{\"o}bian pitfalls have the potential to allow agents to be either arbitrarily manipulated or kept from taking certain actions by trolls. 
Aside from these, the remaining Achilles Heels discussed in Section \ref{survey} are EDT, CDT, updateful decision theory, Sleeping Beauty assumptions, divergent temporal models, and aversion to subjective priors. 
Each of which causes vulnerabilities by making a system prone to a Dutch Book or, more generally, just making bad bets. 
For this reason, so long as failure in one of these situations can be connected to something desirable for containment such as the forfeiture of instrumental resources \cite{hutter2004universal}, then they could also be useful for containment. 
However, there are two broad issues with using bad-bet-based methods for containment.

First, they would require adherence to the rules of some sort of process (\emph{e.g.}, an agent in a Newcombian game can't steal the \$1,000,000).
While it can be possible to compel certain systems to follow certain rules, agents of arbitrary power cannot be expected to follow rules imposed on them in general. 
One potential solution would be to build an incentive to follow rules of such a process into an agent's reward function.
Although this would be a significant specification problem in and of itself, it would be strictly simpler than the broader problem of aligning AI systems with human-compatible goals overall.

Second, a containment strategy which operates based on a bad-bet mechanism will not be unique.
Something similar could be accomplished by a strategy which is not adversarial to an agent's reward function. 
In each of these dilemmas, an agent with an Achilles Heel is motivated by prospects for reward, so the same type of behavior could be elicited via simple bribes of reward. 
For example instead of exploiting an agent to forfeit instrumentally valuable resources in order to chase the folly of an infinite reward in a procrastination paradox, the same behavior could be more straightforwardly elicited by actually giving the agent a reward.

\section{Discussion} \label{discussion}

\begin{table}[]
    \centering
    \small
    \begin{tabular}{|l|l|l|l|l|l|}
        \hline
        & \textbf{Implantation} & \textbf{Stability} & \textbf{Impairment} & \textbf{Subtlety} & \textbf{Containment}\\   \hline
        \textbf{Corrigibility} & Indifference  & --- & Intervention & Usefully non-subtle & Potential\\   \hline
        \textbf{EDT} & Factual info & --- & Bad bets & --- & --- \\  \hline
        \textbf{CDT} & Factual/ctr.factual info & --- & Bad bets & --- & ---\\   \hline
        \textbf{Updateful-DT} & Updatefulness & --- & Bad bets & --- & ---\\   \hline
        \textbf{Sim. Belief} & Pedagogy & Instability risk & Passive & Usefully non-subtle & Potential\\   \hline
        \textbf{Sleeping Beauty} & Pedagogy (SSA/SIA) & Instability risk & Bad bets & Side-effects & --- \\  \hline
        \textbf{Divergence} & Temporal-recursiveness & --- & Bad bets & --- & ---\\   \hline
        \textbf{No Subj. Priors} & Pedagogy & Instability risk & Bad bets & Non-subtlety risk & ---  \\ \hline
        \textbf{L{\"o}bian} & Pedagogy & Instability risk & Sp.proofs/trolls & --- & Potential \\ \hline
    \end{tabular}
    \caption{\textbf{The Achilles Heels presented in this paper with select comments on the criteria from Section \ref{intro} and containment prospects.} For several Achilles Heels, a particular potential for instability or non-subtlety is noted but does not indicate that others are immune to instability or spillover. None are.}
    \label{tab:comments}
\end{table}

This work introduces the \emph{Achilles Heel hypothesis} concerning how highly-effective goal-oriented agents (even ones who are potentially superintelligent) may nonetheless possess stable delusions which can cause failure in adversarial situations.
A number of these potential Achilles Heels are outlined alongside methods by which they can be implanted.
Table \ref{tab:comments} summarizes the Achilles Heels discussed here and includes brief notes related to implantation, stability, impairment, subtlety, and prospects for containment.
It is possible that novel Achilles Heels exist other than the ones presented here, and uncovering additional ones may be a promising direction for further work.
In general, humans finding any problem difficult or paradoxical is a potential sign that an AGI/ASI system might as well. 
Challenging the Achilles Heel hypothesis may also lead to new insights.

A limitation of this work is its focus on understanding sufficiency criteria for when a system will have an Achilles Heel. 
Given how AI systems will not generally adhere to a simple, coherent set of decision theoretic principles, proving necessity criteria will involve additional assumptions.
Nonetheless, these may be valuable for developing theoretical assurances for when a system will \emph{not} develop one of these weaknesses. 
Formulating additional strategies for how certain decision theoretic tendencies can be implanted into systems would be valuable for assurance.
Progress in symbol grounding and related questions involving theory of mind for AI systems will be helpful toward this goal.

This author believes that understanding how advanced AI systems may develop Achilles Heels and the various dilemmas which may exploit them will be valuable for modeling these systems and thinking about strategies for containing them and making them more robust.
A thorough understanding of Achilles Heels will be best paired with well-developed specification and transparency measures for ensuring the aligned development of advanced AI. 
Although modeling these systems remains a difficult challenge, this work makes progress toward understanding novel failure modes and why even superintelligent systems may possess them.

\section*{Acknowledgements} \label{acknowledgements}

This work has benefitted from the feedback of Rohil Badkundri, Luke Bailey, Michael Dennis, Daniel Filan, Adam Gleave, Erik Jenner, Koen Holtman, Pranav Misra, Rohin Shah, Johannes Treutlin, and Alex Turner. 
Caspar Oesterheld in particular provided valuable insights and discussion.

\small
\bibliographystyle{plain}
\bibliography{bibliography.bib}

\appendix

\newpage

\section{Appendix} \label{appendix}

\subsection{``Explain it to me like I'm a highschooler.''}

Much progress is being made in AI and related fields, but there is a communication gap between researchers and the public which often serves as a barrier to the spread of information. This paper is primarily meant for people who are familiar with AI and decision theory, but others may benefit from this brief section dedicated to explaining key concepts free of jargon. 

\bigskip

\noindent Recently, we have witnessed immense progress in AI including systems which, much like a human, can learn to accomplish goals in an environment. These agents are powerful problem-solvers, but they also pose hazards. In fact, there are numerous examples of cases in which learning systems have developed unexpected solutions to problems they are given \cite{lehman2018surprising}.

What happens if and when we develop machines that are as smart as or smarter than humans? In a narrow sense, this has already happened. For example, AI systems can already beat any human at the board games of Chess, Go, and Shogi \cite{schrittwieser2020mastering}. There is nothing to worry about here, but should AI with broad intelligence at this level be created, it might begin a runaway process of achieving its goals in an unintended way. This could lead to unpredictable behavior and may even pose catastrophic threats. Consider an example from \cite{bostrom2017superintelligence}. Imagine a superintelligent AI agent which is meant to optimize paperclip production in a factory. If it is given goals which incentivize it to maximize the number of paperclips that are produced, it could very well try to take over the world with the sole goal of producing as many as possible. The amount of power that a superhuman intelligence could wield is an unknown unknown, and these types of scenarios are a concern.

But regardless of whether highly intelligent AI poses risks on a small scale or an intergalactic one, it remains important to understand how these systems will ``think'' and in what ways they may fail. This paper introduces the concept of ``Achilles Heels'' for AI systems. The idea behind an Achilles Heel is to be a subtle and stable weakness which will not cause bad decisions to be made in normal situations but can cause errors in certain niche ones. As is argued here, even a superintelligent agent which is excellent at achieving its goals in normal cases could make obvious mistakes when presented with these situations. Understanding these Achilles Heels is important for two reasons: (1) the prevention of unintended failures and (2) introducing reliable ``failure'' modes into systems that can help us to maintain control over them where possible. 

By definition, it is hard to think of ways that a system with intelligence at or above the human level may fail because if a human can see that a choice is bad, then such a system ought to as well. However, we can look for clues from tricky problems that often stump humans: paradoxes. The word “paradox” is often used to refer to problems that are merely confusing or counterintuitive such as the Friendship Paradox: that the average person has fewer friends than their average friend does. Other times, ``paradox'' refers to problems of vagueness or self-contradiction such as the Liar’s Paradox: ``This sentence is a lie.'' But in other cases, paradoxes are genuine theoretical problems which teach us that some subtle aspect of how one might think about identity, causation, infinity, or inference is mistaken and that a mode of reasoning which might work great 99.9\% of the time can egregiously fail that other 0.1\%. 

As a simple example, Evidential Decision theory (EDT) is a decision-making philosophy which says to take the action that is evidentially consistent with the best outcome. In other words, EDT says to act in the way that it would be the best news to find out you had acted. Normally, this is an excellent rule which will recommend making optimal choices. But in some cases, it can make very bad recommendations. Imagine that there is a genetic mutation possessed by some humans which has two effects: a very painful disorder and a love for bananas. Suppose that you are unaware of your genotype but that you want to eat a banana. Should you eat it? Of course! There is no causal relationship between eating the banana and whether or not you have the disorder. Yet EDT would say not to eat the banana merely because it is correlated with having the disease!

The bulk of this paper is dedicated to surveying work involving paradoxes and dilemmas and arguing that a number of them are likely to be subtle and stable Achilles Heels and presenting means by which these weaknesses could be implanted into AI systems. 

\begin{itemize}
    \item \textit{Corrigibility:} One desirable Achilles Heel for an advanced AI agent to have is the ability to be reliably turned off or altered if it is observed to be misbehaving. This is a simple goal, but it is complicated by the fact that for the same reason you would not want to die or become a zombie, a goal-oriented AI system would not want to be turned off or altered. Several techniques for making systems amenable to corrections are discussed. 
    \item \textit{Evidential and Causal Decision Theories:} Evidential Decision Theory and a related philosophy known as Causal Decision Theory both usually make the right decisions but can prescribe problematic actions in cases in which correlations are non-causal and/or an agent interacts with a model of itself. 
    \item \textit{Anthropic Assumptions:} Under certain philosophical assumptions, one’s own existence can be relevant evidence involving questions about a process involved with how they came into existence. An agent with the wrong assumptions can be tricked into making bad bets in some situations.
    \item \textit{Divergent Temporal Models (a.k.a. Misconceptions about Infinity):} A number of paradoxes involve the concept of infinity and how things can go wrong when one treats it as a number rather than a limit or refuses to put bounds on how high they are willing to count. 
    \item \textit{Aversion to Subjective Priors:} Sometimes, if one acts as if they are completely ignorant about an unknown value and unwilling to assign any sort of prior beliefs to it, they can make decisions which allow them to be systematically exploited. 
    \item \textit{L{\"o}bian Pitfalls:} An important theorem pertaining to logic known as L{\"o}b's Theorem can be used to show that agents can be vulnerable to arbitrary manipulation or making decisions that are too risk-averse if they have the wrong views about certain hypothetical situations. 
\end{itemize}

This paper surveys, discusses, and augments a great deal of prior work in decision theory on these tricky dilemmas and unifies them under this new Achilles Heel framework. Hopefully, this and related work will contribute to a better understanding of these problems and how they relate to building more robust and controllable AI systems. This paper might also serve as a useful reference for those who are intrinsically interested in rational decision making.

\end{document}